%% file: main.tex
\documentclass[a4paper,conference]{IEEEtran}

\ifCLASSINFOpdf
   \usepackage[pdftex]{graphicx}
\else
   \usepackage[dvips]{graphicx}
\fi
\usepackage{amsmath}
\usepackage{url}

\usepackage{hyperref}
\hypersetup{
    colorlinks=true, 
    linkcolor=blue,
    filecolor=magenta,      
    urlcolor=cyan,
}

\input{preamble}

\begin{document}
%
\title{Supervised Domain Adaptation \\ using Graph Embedding}


\author{
    \IEEEauthorblockN{
        Lukas Hedegaard Morsing\textsuperscript{*},
        Omar Ali Sheikh-Omar\textsuperscript{*}, and 
        Alexandros Iosifidis
    }
    \IEEEauthorblockA{
        Department of Engineering,
        Aarhus University, Denmark\\
        \href{mailto:lh@eng.au.dk}{lh@eng.au.dk},
        \href{mailto:sheikhomar@mailbox.org}{sheikhomar@mailbox.org},
        \href{mailto:ai@eng.au.dk}{ai@eng.au.dk}
    }
}


\maketitle

\begingroup\renewcommand\thefootnote{*}
\footnotetext{Equal contribution}
\endgroup

\input{content/00-abstract}


%
\IEEEpeerreviewmaketitle

\input{content/01-introduction}
\input{content/02-related-works}
\input{content/03-dage}

\input{content/04-experiments}

\input{content/05-conclusion}
\input{content/06-acknowledgement}






\bibliography{references}

\end{document}

%% file: preamble.tex
\usepackage[capitalise]{cleveref}

\usepackage{booktabs}
\usepackage{mathrsfs}
\usepackage{bm,bbm}
\usepackage{amsfonts}

\usepackage{subcaption}
\usepackage{scrextend}
\usepackage{cancel}

\usepackage[numbers,sort]{natbib}

\newcommand{\halffigwidth}{0.22\textwidth} 

\usepackage{ifthen} 
\newcommand{\blind}{1=0} 

\renewcommand{\vector}[1]{\ensuremath{\bm{\lowercase{#1}}}}
\renewcommand{\matrix}[1]{\ensuremath{\bm{\mathbf{\uppercase{#1}}}}}

\newcommand{\vx}{\vector{x}}
\newcommand{\vy}{\vector{y}}
\newcommand{\vz}{\vector{z}}

\newcommand{\mX}{\matrix{x}}
\newcommand{\mW}{\matrix{W}}
\newcommand{\mD}{\matrix{D}}
\newcommand{\mB}{\matrix{B}}
\newcommand{\mL}{\matrix{L}}
\newcommand{\Domain}{\mathscr{D}} 
\newcommand{\Task}{\mathscr{T}} 
\newcommand{\concept}[1]{\ensuremath{\mathcal{\uppercase{#1}}}}
\newcommand{\set}[1]{\ensuremath{\mathbb{\uppercase{#1}}}}
\newcommand{\norm}[1]{\ensuremath{\left\|#1 \right\|}}

\DeclareMathOperator*{\argmin}{argmin} 

\newcommand{\Src}{\mathcal{S}} 
\newcommand{\Tgt}{\mathcal{T}} 

\newcommand{\tr}{\text{Tr}}

%% file: content/00-abstract.tex
\begin{abstract}
Getting deep convolutional neural networks to perform well requires a large amount of training data. When the available labelled data is small, it is often beneficial to use transfer learning to leverage a related larger dataset (source) in order to improve the performance on the small dataset (target). 
Among the transfer learning approaches, domain adaptation methods assume that distributions between the two domains are shifted and attempt to realign them. In this paper, we consider the domain adaptation problem from the perspective of multi-view graph embedding and dimensionality reduction. Instead of solving the generalised eigenvalue problem to perform the embedding, we formulate the graph-preserving criterion as a loss in the neural network and learn a domain-invariant feature transformation in an end-to-end fashion. We show that the proposed approach leads to a powerful Domain Adaptation framework which generalises the prior methods CCSA and \textit{d}-SNE, and enables simple and effective loss designs; an LDA-inspired instantiation of the framework leads to performance on par with the state-of-the-art on the most widely used Domain Adaptation benchmarks, Office31 and MNIST to USPS datasets. 

\end{abstract}

%% file: content/01-introduction.tex
\section{Introduction} \label{sec:introduction}

It has been repeatedly shown that deep convolutional neural networks (CNNs) perform remarkably well for various supervised tasks in computer vision
\cite{krizhevsky2012imagenet, redmon2015look}.
However, the usefulness of these deep learning methods for real-world applications are limited when the training data is scarce. This is because the convolutional filters require huge amounts of training examples to learn and extract useful features from images~\cite{oquab2014learning}. Moreover, the fully connected layers of deep CNNs easily overfit small training data because their large capacity allows them to memorise the training examples~\cite{arpit2017closer}.

Inspired by how humans tackle new learning tasks by relying on prior knowledge from related domains, \emph{transfer learning} aims to alleviate the issue of data scarcity. 
The main idea is to leverage information contained in a related larger dataset in order to improve the performance on a smaller dataset. We denote these by \emph{source domain} $\Domain_\Src$ and \emph{target domain} $\Domain_\Tgt$ respectively.

More formally, a domain $\Domain=\{\concept{X}, p(\mX)\}$ can be defined as a feature space $\concept{X}$ and a marginal probability distribution $p(\mX)$ of samples in that space $\mX = \{\vx_1, \dots, \vx_N\} \in \concept{X}$. 
We want to learn an objective predictive function $f(\cdot)$ that predicts a label $y_i$ of a label space $\concept{Y}$ from the corresponding sample $\vx_i$. This constitutes a task, $\Task=\{\concept{Y}, f(\cdot)\}$~\cite{pan2010survey}.
Transfer learning is the attempt of learning a target task $\Task_\concept{T}$ from both source and target data $\mX_\Src$ and $\mX_\Tgt$ when there is a discrepancy in either the domains ($\Domain_\Src \neq \Domain_\Tgt$) or tasks ($\Task_\Src \neq \Task_\Tgt$).
A source domain and a target domain are said to be different if the feature spaces are not the same ($\concept{X}_\Src \neq \concept{X}_\Tgt$) or their marginal probability distributions are unequal, i.e. $p(\mX_\Src) \neq p(\mX_\Tgt)$. In such cases, an effective way to learn the target task is to explicitly map the data to a common, domain invariant, representation. This learning tactic is called \emph{domain adaptation} (DA).
When the domain difference is one of distribution difference ($p(\mX_\Src) \neq p(\mX_\Tgt)$), sometimes referred to as \emph{domain shift} or \emph{covariate shift}, it is called \emph{homogenous domain adaptation}~\cite{wang2018deep}. This is the most prevalent type domain adaptation and also the focus of our effort.

\begin{figure}[!b]
    \centering
    \includegraphics[width=0.9\linewidth]{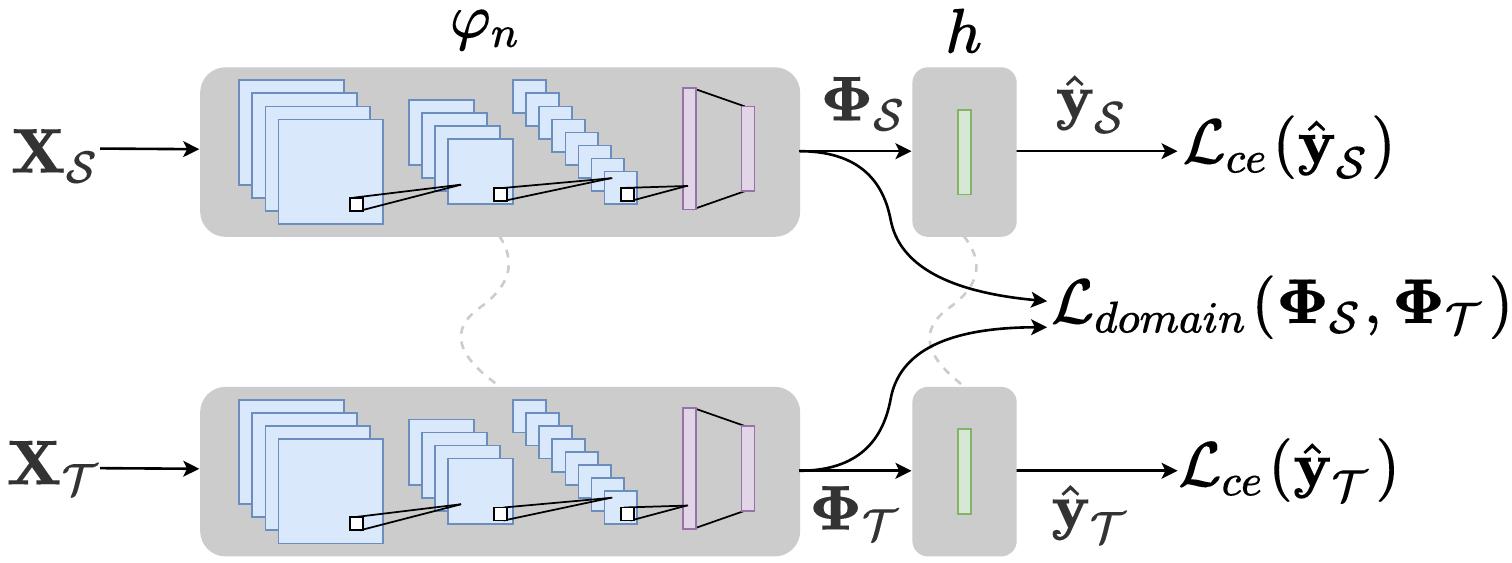}
    \caption{Siamese network architecture for domain adaptation. Samples from the source and target domains are introduced to an embedding function $\varphi$ consisting of a convolutional neural network. This produces a lower dimensional embedding for which an inter-domain loss can be computed. A prediction function $h$ uses these embeddings to predict corresponding labels, for which a categorical cross-entropy loss is applied to evaluate each prediction at training time. Network updates are performed jointly from cross-entropy and domain adaptation losses, and network parameters are shared across both streams. }
    \label{fig:two-stream-architecture}
\end{figure}

In this line of research, it is usually assumed that the source and target label spaces are the same.
Depending on the availability of labels in the target domain, DA methods are categorised further into three buckets~\cite{wang2018deep}. Supervised DA methods assume that the available target data is labelled albeit small. In the semi-supervised setting, in addition to a small amount of labelled target data, a larger amount of unlabelled target data is available. If only unlabelled target data is available, is it denoted unsupervised. 
The reader should note that many different and occasionally conflicting definitions of semi-supervised and unsupervised DA were presented in
\cite{pan2010survey, wang2018deep, cook2013, feuz2015, daume2009, chatto2011, gong2012}. 

When training data (e.g. images) is scarce, it might be straightforward to acquire a large number of images by scraping the web. However, the process of labelling these images is time-consuming and costly because of the manual human labour associated with the data annotation process. It is, therefore, of no surprise that the majority of domain adaptation methods in the literature focus on the unsupervised settings. 
However, in the case where the number of samples per class is very small, supervised domain adaptation approaches outperform the unsupervised methods~\cite{motiian2017ccsa}.

A popular approach employed in the few-shot supervised DA methods is to learn a feature transformation that maps same-class samples close together in a common latent space while pushing samples with different labels farther apart~\cite{motiian2017ccsa}, \cite{xu2019dsne}. This can be done by utilising a Siamese network architecture~\cite{bromley1993} (see \cref{fig:two-stream-architecture}) and feeding pairs of samples from both domains as input. A loss function is defined at the output of the feature extractor of a deep CNN to encourage it to generate domain-invariant features. 

In our work, we view domain adaptation from the vantage point of dimensionality reduction, and introduce a generic framework for the supervised DA problem using graph embedding. Inspired by Yan et al.~\cite{yan2006mfa}, we create two graphs. An \emph{intrinsic graph} allows us to encode within-class compactness criteria and a \emph{penalty graph} models between-class separability criteria. By connecting samples across domains, we can enforce desired properties such as domain invariance.   
One of the key advantages of this approach is that domain knowledge and design decisions can be easily encoded as rules of graph construction. We show how to construct an effective graph for domain adaptation inspired by Linear Discriminant Analysis.

Traditionally, graph embedding methods are optimised by solving the generalised eigenvalue problem. Recently, an unsupervised DA method solving the eigenproblem was proposed~\cite{chen2019gef}. By contrast, our method uses the Siamese network architecture and is optimised in an end-to-end fashion using Backpropagation. Similar to other representation learning-based DA methods, we want to learn a non-linear feature transformation which generates domain-invariant and semantically meaningful features.
Our contributions are:
\begin{enumerate}
    \item We propose a graph-based domain adaptation method which performs on par with the current state-of-the-art. Graphs are conceptually easy to understand and it is straightforward to tailor them to specific problems.
    \item We show how graph embedding-based domain adaptation methods can be optimised in an end-to-end fashion in a deep neural network instead of solving the generalised eigenvalue problem. Our method can scale when the number of samples in the source domains is very large.
\end{enumerate}

The rest of the paper is structured as follows. In \cref{sec:related-work}, we discuss other existing methods in the supervised DA setting. \cref{sec:dage} details how we derive a general framework by posing the domain adaptation problem as dimensionality reduction. Moreover, we present a configuration of the proposed Domain Adaptation using Graph Embedding (DAGE) framework inspired by Linear Discriminant Analysis. We evaluate the algorithm on few-shot domain adaptation tasks using two canonical image datasets Office31 and MNIST to USPS in \cref{sec:experiments}. Finally, in \cref{sec:conclusion} we summarise and provide suggestions for future research.

%% file: content/02-related-works.tex
\section{Related Works} \label{sec:related-work}





In the supervised DA setting, two recent works focus specifically on the problem of few-shot learning. Classification and Contrastive Semantic Alignment (CCSA)~\cite{motiian2017ccsa} uses the idea of contrastive loss introduced by Hadsell  et  al. \cite{hadsell2006dimensionality} to pull together same-label samples in an embedding space and push away samples with different labels if they are within a given margin $m$. Sample pairs from source and target domains are fed as input to a Siamese deep neural network alongside an indication of whether the labels of the sample pair are the same ($\alpha=1$) or different ($\alpha=0$). Using the Euclidean distance of next-to-last layer features, $d(\vx_\Src, \vx_\Tgt)=\norm{\varphi_n(\vx_\Src), \varphi_n(\vx_\Tgt)}_2$, the CSA domain adaptation loss is computed over the sum of pairs. This criterion is minimised jointly with the classification loss using gradient descent.
\begin{align}\label{eq:csa_loss_generic}
    \concept{L}_{\text{CSA}} 
    =& 
    \sum_{\vx_\Src \in \Domain_\Src}
    \sum_{\vx_\Tgt \in \Domain_\Tgt}
    \alpha \ \frac{1}{2} d(\vx_\Src, \vx_\Tgt)^2 
    \nonumber
    \\
    &+ (1-\alpha) \ \frac{1}{2} 
    \left( 
        \max\{0, m -  d(\vx_\Src, \vx_\Tgt)\}
    \right)^2
\end{align}
Domain Adaptation using Stochastic Neighborhood Embedding ($d$-SNE) uses the same training setup as CCSA. However, instead of minimising the distances of all pairs with the same label, the $d$-SNE loss~\cite{xu2019dsne} focuses on reducing the distance of the same-class pairs with the largest distance, and maximising the distance of the different class pairs that are closest on one another. For each batch of data, the $d$-SNE loss is defined as
\begin{align} \label{eq:dsne_loss_relax}
\concept{L}_{\text{d-SNE}}
    =& \sum_{j \in \Domain_\Tgt}
    \sup_{\vx \in \Domain_{\Src}^{c}}{\left\{ a \mid a \in d(\vx, \vx_{\Tgt}^{j}) \right\}}
    \nonumber
    \\
    &- \inf_{\vx \in \Domain_{\Src}^{\cancel{c}}}{\left\{ b \mid b \in d(\vx, \vx_{\Tgt}^{j}) \right\}}
    \textrm{, for } c = y_j
\end{align}

Our work also uses the two-stream architecture and is trained similarly to CCSA and $d$-SNE. 
We, however, view the deep neural network as a function performing dimensionality reduction, and can utilise well-tested techniques from graph embedding to learn a non-linear feature transformation which is both domain-invariant and semantically meaningful.

Closest to our approach is the Graph Embedding Framework for Maximum Mean Discrepancy-Based Domain Adaptation Algorithm (GEF)~\cite{chen2019gef}. Their work is unsupervised and focuses on Maximum Mean Discrepancy-based DA methods. The GEF algorithm is less appropriate for larger datasets as it solves the generalised eigenvalue problem in each iteration to generate pseudo labels. 
Another related unsupervised method is Scatter Component Analysis~\cite{ghifary2017}, which uses the kernel trick to project samples to Hilbert space and solves domain adaptation analytically as a generalised eigenvalue problem. In contrast to these methods, our proposed method integrates the graph embedding objective in a deep neural network and is trained using Backpropagation.


%% file: content/03-dage.tex
\section{Methodology}
\label{sec:dage}

In this section, we present the proposed framework for domain adaptation. We start with a description of graph embedding. 

\subsection{Graph Embedding}
Graph Embedding is a process of mapping the vertices of a graph into low-dimensional vectors which preserve the relationship structure inherent in the graph
\cite{goyal2018graph, yan2006mfa}.
It is a technique in representation learning which aims to distil useful information from high-dimensional data.

Let $\mX = [\vx_1, \cdots, \vx_N] \in \set{R}^{D\times N}$ be matrix of the training samples and $\vz = [z_1, \cdots, z_N]^\top \in \set{R}^{1\times N}$ be their corresponding one-dimensional representation. To represent similarity relationships between pairs of samples, we create an undirected graph $G = (\mX, \mW)$ where each column in $\mX$ represents a vertex and $\mW \in \set{R}^{N \times N}$ is the similarity matrix encoding pair-wise similarities between the graph vertices, i.e. the entry $\mW^{(i,j)}$ expresses the similarity between vertices $\vx_i$ and $\vx_j$.

To obtain the optimal embeddings $\vz^*$ while preserving the similarity characteristics defined in the graph $G$, the \emph{graph-preserving criterion}~\cite{yan2006mfa} is optimised:
\begin{align}
  \vz^{*} = \argmin_{
    \vz^\top \mB \vz = c
  }
  \sum_{i \neq j}
  \norm{
    z_i 
    -
    z_j 
  }^2
  \mW^{(i,j)}
  \label{eq:graph_preserving_criterion}
\end{align}
where $c$ is a constant and $\mB \in \set{R}^{N \times N}$ is a constraint matrix used to avoid trivial solutions. \cref{eq:graph_preserving_criterion} can also be formalised in terms of a graph Laplacian:
\begin{align}
  \vz^{*} = \argmin_{
    \vz^\top \mB \vz = c
  }
  \vz^\top \mL \vz
  \label{eq:graph_preserving_criterion_laplacians}
\end{align}
where $\mL = \mD - \mW$ is the Laplacian matrix of $G$ and $\matrix{D}$ is the (diagonal) degree matrix with entries $\matrix{D}^{(i,i)} = \sum_{i\neq j} \mW^{(i,j)}$

The constraint matrix $\mB$ can also be the Laplacian matrix of a penalty graph $G_p = (\mX, \mW_p)$, i.e. $\mB = \mD_p - \mW_p$ where $\matrix{D}_p^{(i,i)} = \sum_{i\neq j} \mW_p^{(i,j)}$. When we design the penalty graph to represent undesirable characteristics to be suppressed, optimising the graph-preserving objective jointly promotes desired similarity characteristics and penalises undesired characteristics. This framework can also be used to express our desire to create domain-invariant embeddings. 

\subsection{Domain Adaptation using Graph Embedding (DAGE)}
In representation-based DA, the goal is to learn a transformation $\varphi(\cdot)$ where same-class features are placed close together in the embedding space (\emph{within-class compactness}) while features with different labels are pushed farther apart (\emph{between-class separability}) irrespective of their originating domain. We express this goal using graph embedding.

Suppose $\mX_\Src \in \set{R}^{D\times N_\Src}$ and $\mX_\Tgt \in \set{R}^{D\times N_\Tgt}$ are two matrices containing the training data from the source and target domains, respectively. Let $
\matrix{\Phi}
=
[
\varphi( \mX_\Src )
, 
\varphi( \mX_\Tgt )
]
\in 
\set{R}^{d \times N  }
$ be a matrix of the feature representations for source and target data, where $N = N_\Src + N_\Tgt$. The within-class compactness objective can be achieved by minimising the expression:
\begin{align}
  \min
  \sum_{
    \substack{i=1}
  }^{N}
  \sum_{
    \substack{j=1}
  }^{N}
  \norm{
    \matrix{\Phi}^{(i)}
    -
    \matrix{\Phi}^{(j)}
  }_2^2
  \matrix{W}^{(i,j)}
  =
  \tr \left(
  \matrix{\Phi}
  \matrix{L}
  \matrix{\Phi}^\top
  \right)
  \label{eq:within_class_compactness_criterion}
\end{align}
where $\matrix{W}$ encodes how the distance between a pair of features should be weighted with non-zero edges for sample-pairs corresponding to different domains.
Similarly, the between-class separability objective can be expressed as:
\begin{align}
  \max
  \sum_{
    \substack{i=1}
  }^{N}
  \sum_{
    \substack{j=1}
  }^{N}
  \norm{
    \matrix{\Phi}^{(i)}
    -
    \matrix{\Phi}^{(j)}
  }_2^2
  \matrix{W}_p^{(i,j)}
  =
  \tr \left(
  \matrix{\Phi}
  \matrix{B}
  \matrix{\Phi}^\top
  \right)
  \label{eq:between_class_separability_criterion}
\end{align}

Combining \cref{eq:within_class_compactness_criterion} and \cref{eq:between_class_separability_criterion}, we formulate the domain adaptation objective as a trace-ratio minimisation criterion:
\begin{align}
  \varphi^{*}
  &=
  \argmin_{\theta_{\varphi}} 
  \frac{
    \tr
    \left(
    \matrix{\Phi}
    \matrix{L}
    \matrix{\Phi}^\top
    \right)
  }{
    \tr
    \left(
    \matrix{\Phi}
    \matrix{B}
    \matrix{\Phi}^\top
    \right)
  }
  \label{eq:dage_criterion}
\end{align}
where $\theta_{\varphi}$ are the parameters of the embedding function.

Traditionally, graph embedding methods optimise either a linear projection $\varphi(\mX)=\vector{a}^{\top}\mX$ or exploit the Representer Theorem to form a kernelised embedding function $\varphi(\mX)=\vector{a}^{\top}\matrix{K}$. Here, $\matrix{K}$ is the (kernel) Gram matrix with $\matrix{K}^{(i,j)} = k(\vx_i, \vx_j)$ and $k(\cdot,\cdot)$ is a nonlinear function evaluating the similarity between its entries~\cite{yan2006mfa}. The problem is optimised by solving the generalised eigenvalue problem.

We opt for a third choice:
A CNN can be seen as a function $f(\vx)=h(\varphi_n(\vx))$ where  $\varphi_n : \concept{X} \rightarrow \concept{Z}$ is a neural network used as a feature extractor which maps input data to a lower-dimensional subspace $\concept{Z}$ and $h : \concept{Z} \rightarrow \concept{Y}$ is the classifier which maps features to the label space $\concept{Y}$. 
Note that by using this choice, our approach is also able to optimise the data representation $\varphi_n(\vx)$ based on the classification criterion as well.
In practice, we use a Siamese network architecture with shared weights across both streams as shown in \cref{fig:two-stream-architecture}. By simultaneously computing the embeddings for data from source and target domains, we optimise for a graph embedding loss for each batch of data:
\begin{align}
  \concept{L}_{\text{DAGE}}
  &=
  \frac{
    \tr
    \left(
    \begin{bmatrix}
      \varphi_n(\mX_\Src) &
      \varphi_n(\mX_\Tgt)
    \end{bmatrix}
    \matrix{L}
    \begin{bmatrix}
      \varphi_n(\mX_\Src) &
      \varphi_n(\mX_\Tgt)
    \end{bmatrix}^\top
    \right)
  }{
    \tr
    \left(
    \begin{bmatrix}
      \varphi_n(\mX_\Src) &
      \varphi_n(\mX_\Tgt)
    \end{bmatrix}
    \matrix{B}
    \begin{bmatrix}
      \varphi_n(\mX_\Src) &
      \varphi_n(\mX_\Tgt)
    \end{bmatrix}^\top
    \right)
  }
  \label{eq:dage_loss}
\end{align}
%

For $\matrix{\Phi} = \begin{bmatrix}
      \varphi_n(\mX_\Src) &
      \varphi_n(\mX_\Tgt)
    \end{bmatrix} $ the gradient of this expressions is:
\begin{align}
  \nabla_{\matrix{\Phi}} \mathcal{L}_{\text{DAGE}}
  &=
  \frac{
    \tr 
    \left(
      \matrix{\Phi}
      \matrix{L}^\top
      +
      \matrix{\Phi}
      \matrix{L}
    \right)
  }{
      \tr 
      \left( 
        \matrix{\Phi} 
        \matrix{B} 
        \matrix{\Phi}^\top 
      \right)
  }
  -
  \frac{
    \tr 
      \left(
        \matrix{\Phi} 
        \matrix{L}
        \matrix{\Phi}^\top
      \right)
      \left(
        \matrix{\Phi}
        \matrix{B}^{\top}
        +
        \matrix{\Phi}
        \matrix{B}
      \right)
  }{
      \tr 
      \left( 
        \matrix{\Phi} 
        \matrix{B} 
        \matrix{\Phi}^\top 
      \right)^2
  }
\end{align}

Finally, we can train the network end-to-end by jointly minimising the DAGE loss and cross-entropy losses for batches of source and target data:
\begin{align}
  \argmin_{\theta_{\varphi}, \theta_{h}} \:
  \concept{L}_{\text{DAGE}}
  +
  \beta \ \concept{L}_{\text{CE}}^{\Src}
  +
  \gamma \ \concept{L}_{\text{CE}}^{\Tgt}
  \label{eq:tot_loss}
\end{align}
where $\beta$ and $\gamma$ weight the source and target cross-entropies. For $M$ samples and $K$ classes the cross-entropy is given as:
\begin{align}
    \concept{L}_{\text{CE}}(\vy,\hat{\vy})
    =
    - \sum_{i=1}^M \sum_{k=1}^K \vy_{ik} \ln{\hat{\vy}_{ik}}
\end{align}
where $\vy_{ik}$ is 1 if sample $i$ belongs to class $k$, and is 0 otherwise, and $\hat{\vy}_{ik}$ is the predicted probability that $\vx_i$ belongs to class $k$.

The observant reader may notice that the optimisation target of \cref{eq:tot_loss} is very similar to those of CCSA and $d$-SNE. This is by design. In fact, the framework laid out here generalises CCSA and $d$-SNE as graph embeddings, which can be easily shown by formulating \cref{eq:csa_loss_generic} and \cref{eq:dsne_loss_relax} in terms of the weight matrices that give rise to the intrinsic and penalty graphs of a graph embedding when the equivalent trace-difference~\cite{jia2009trace} formulation is used.


\subsection{DAGE-LDA}
Many different domain adaptation formulations can be achieved based on the choice of encoding rules for the weight matrices associated with the intrinsic and penalty graphs.
One such rule is inspired by Linear Discriminant Analysis (LDA).
\begin{figure}
    \centering
    \begin{subfigure}[b]{\halffigwidth}
        \includegraphics[width=\textwidth]{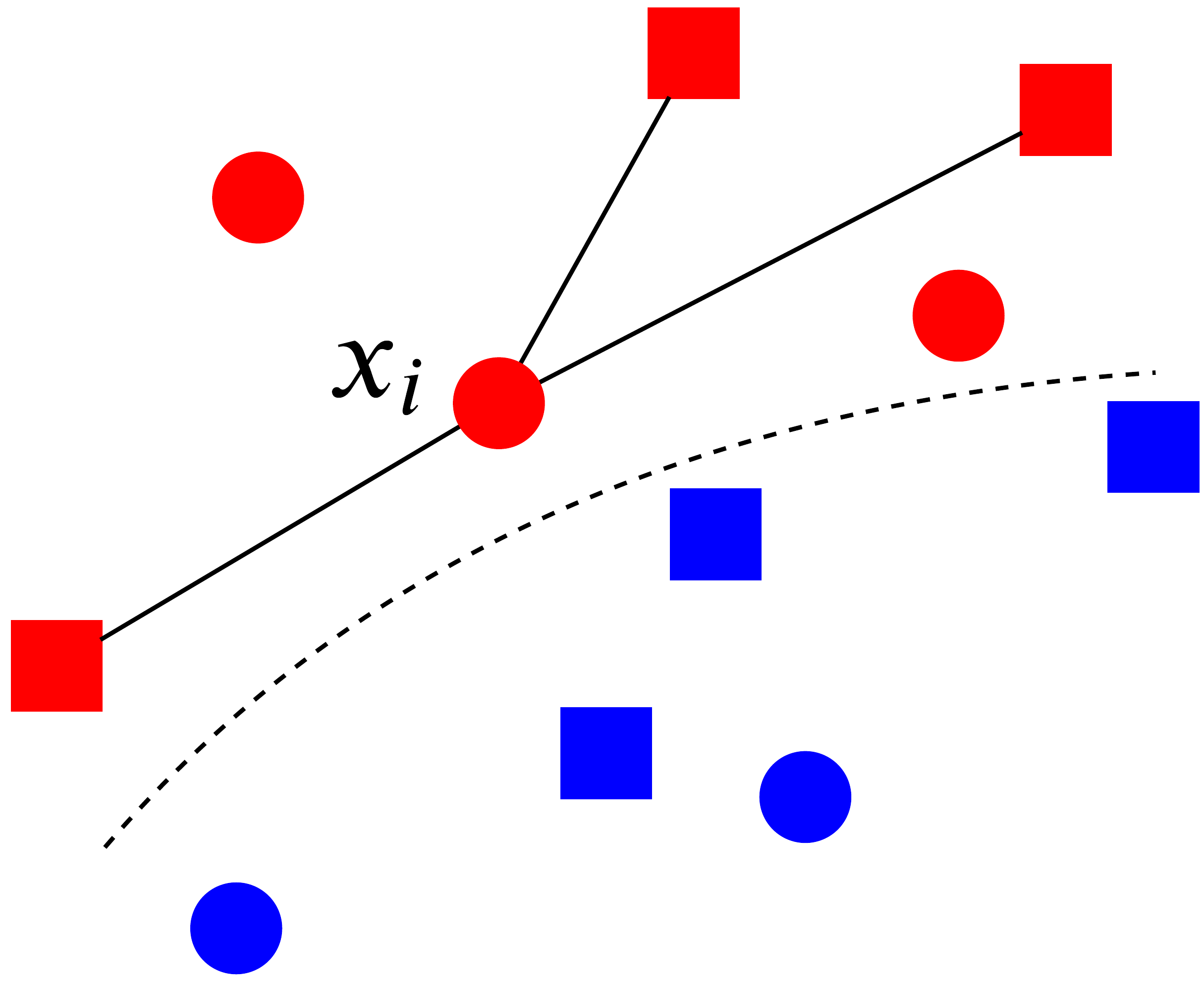}
		\caption{Intrinsic graph.}
		\label{fig:dage-lda-within}
	\end{subfigure}
	\qquad
	\begin{subfigure}[b]{\halffigwidth}
	    \includegraphics[width=\textwidth]{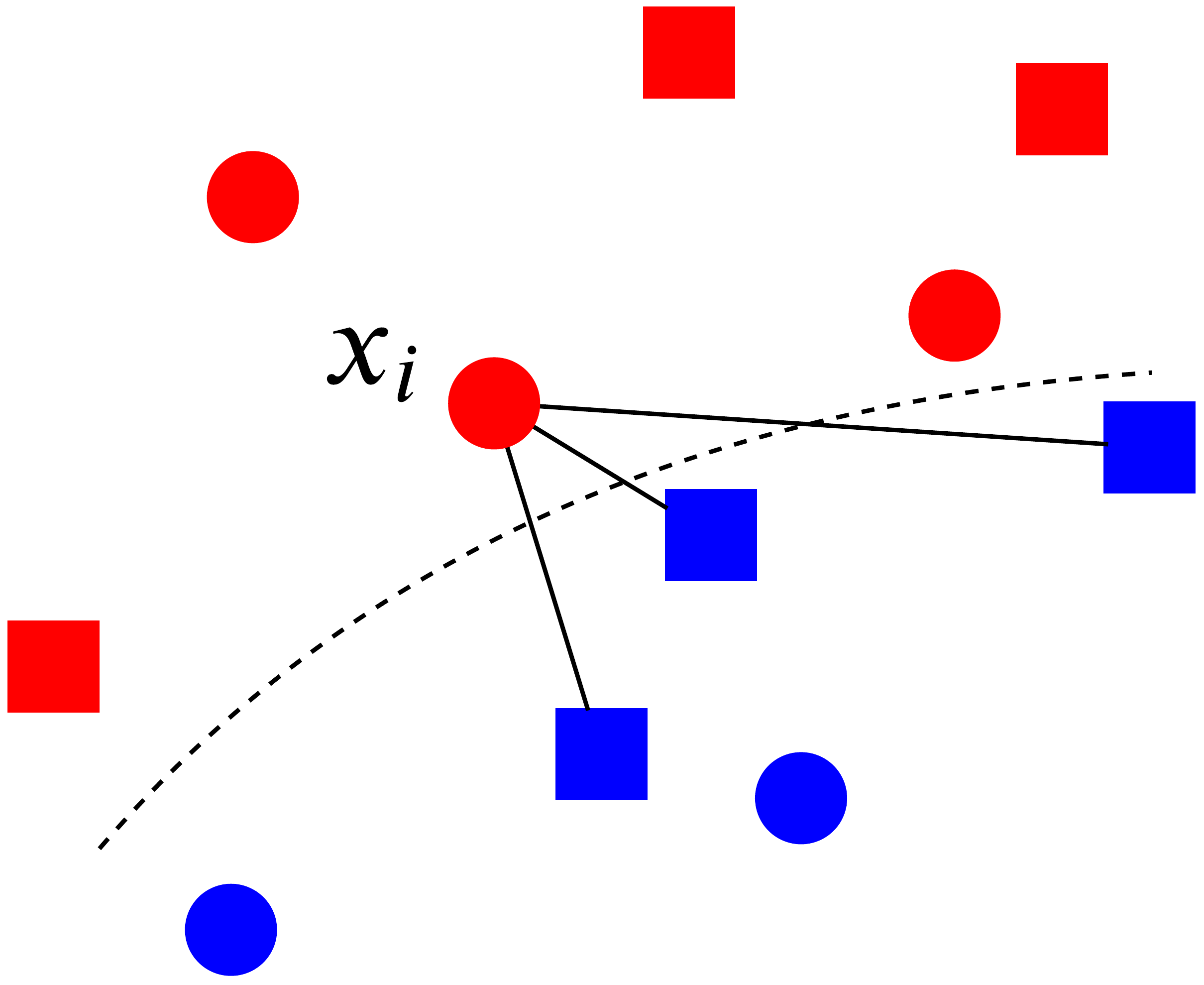}
		\caption{Penalty graph.}
		\label{fig:dage-lda-between}
	\end{subfigure}
    \caption{DAGE-LDA edges in the intrinsic and penalty graphs for a sample $\vx_i$ depicted in an Euclidean space. The shapes mark different domains whereas the colours denote different classes. The dashed line indicates a class boundary.}
    \label{fig:dage-lda-graphs}
\end{figure}
By adopting the simple rule that samples from different domains but of the same class have an edge in the intrinsic graph, we encourage the network to learn a feature space that achieves same-class compactness regardless of the originating domain:
\begin{align}
    \label{eq:dage-lda-w}
    \matrix{W}^{(i,j)} &= 
        \begin{cases}
            1, & \text{if } y_i = y_j \text{ and } \Domain_i \neq \Domain_j \\
            0, & \text{otherwise } \\
        \end{cases} 
\end{align}
%
In practice, we found the removal of within-domain edges to work slightly better, albeit with a negligible difference. 

The corresponding penalty weight matrix is
\begin{align}
\label{eq:dage-lda-w-p}
    \matrix{W}_{p}^{(i,j)} &= 
        \begin{cases}
            1, & y_i \neq y_j \text{ and } \Domain_i \neq \Domain_j \\
            0, & \text{otherwise } \\
        \end{cases} 
\end{align}
which produces the inverse across-domain edges as compared to the intrinsic graph.
A depiction of the source an target graphs for a small batch of data is shown in \cref{fig:dage-lda-graphs}

From \cref{eq:dage_loss} it can be seen that if the samples forming the mini-batch all have the same label, the denominator formed by using \cref{eq:dage-lda-w-p} becomes zero. We solve this issue by making sure that batches of training data are shuffled and include samples from multiple classes. Alternatively, one can use a regularised version of the denominator, i.e. by adding a small positive value $\epsilon > 0$.

%% file: content/04-experiments.tex
\section{Experiments \& Results} \label{sec:experiments}

We evaluated DAGE-LDA as well as the recent state of the art methods CCSA~\cite{motiian2017ccsa} and $d$-SNE~\cite{xu2019dsne} on the two image dataset collections typically used for benchmarking domain adaptation methods~\footnote{\ifthenelse{\blind}{A link to the source code will be available upon acceptance}{Source code is available at \url{https://github.com/lukashedegaard/dage}}}.

\begin{figure}[t]
    \centering
    \includegraphics[width=0.9\linewidth]{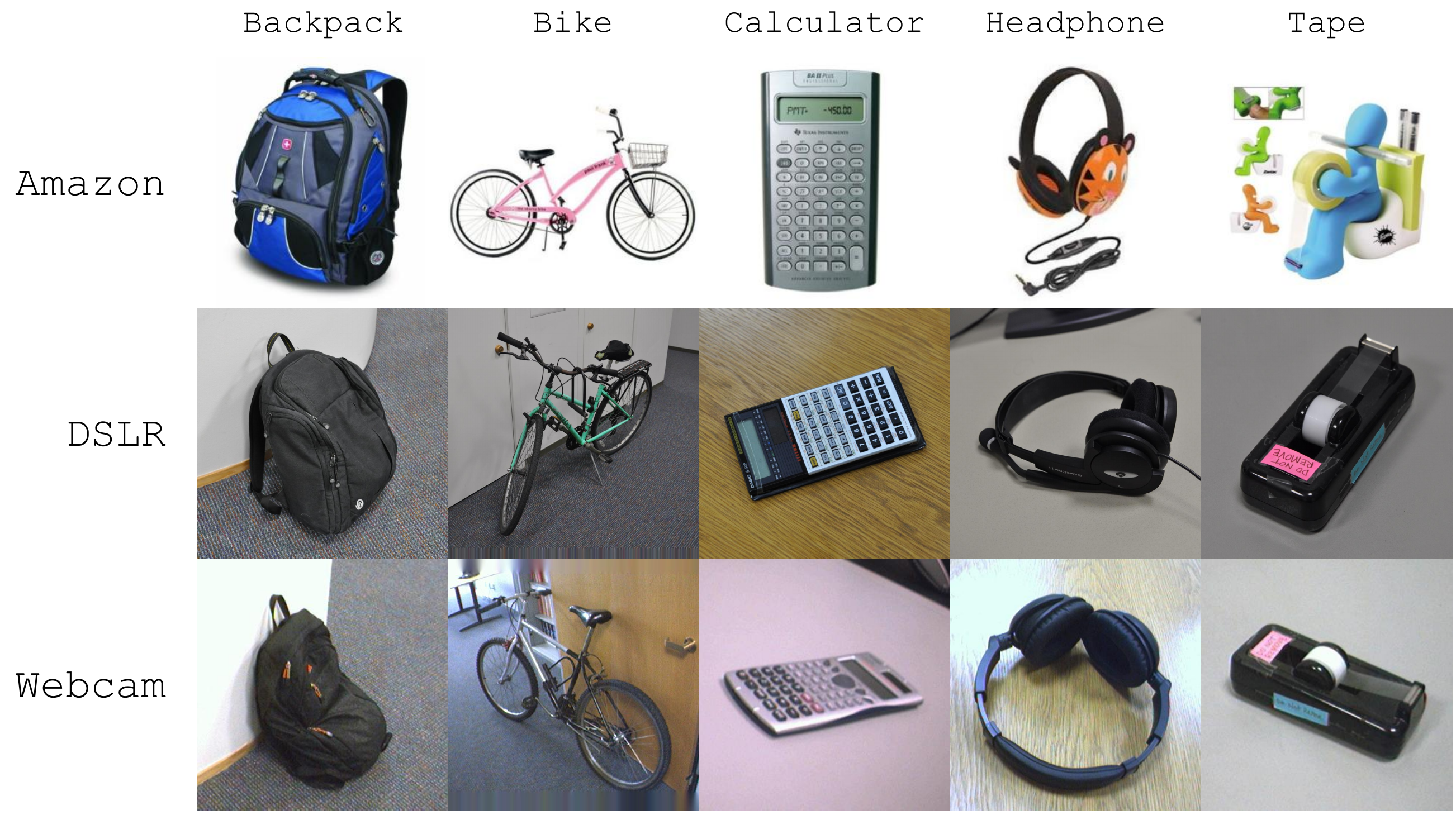}
    \caption{Samples from the Office dataset.}
    \label{fig:office31-examples}
\end{figure}

The Office31~\cite{saenko2010adapting} is a canonical dataset created for the purpose of studying the effects of domain shift in computer vision. It consists of 31 classes from three visual domains: Amazon ($\concept{A}$), DSLR ($\concept{D}$) and Webcam ($\concept{W}$). 
The 2.817 images in $\concept{A}$ are characterised by having distinct objects centred in the image with white backgrounds. The DSLR domain contains 498 high resolution pictures of objects. Within this domain, each of the 31 classes contain images of five different objects. Each object is photographed from three different viewpoints. In the Webcam domain, 795 images of the same five objects as in the DSLR domain were captured using a low-res webcam. These images are noisy and suffer from colour distortions. A sample of the images in the Office31 is shown in \cref{fig:office31-examples}.

Another common domain adaptation task is from the MNIST~\cite{lecun2010mnist} to USPS~\cite{lecun90handwritten} dataset. Both datasets consist of grayscale images of handwritten digits from 0 to 9. MNIST contains 70,000 $28 \times 28$ images and USPS 11,000 $16 \times 16$ images. 

Following the procedure in the source code of \cite{xu2019dsne}, the data is augmented using rotation, zoom, flipping, hue, saturation, brightness, and contrast when relevant. 
To find the best hyper-parameter values for each of the evaluated methods, we employed Bayesian Optimisation using the Expected Improvement acquisition function~\cite{brochu2010} and the search space summarised in \cref{tab:office31-search-space}.

The reader should note that the goal of our experiments is not to achieve the highest absolute accuracy among the possible network designs. As image classification networks evolve, it is trivially easy to improve accuracy by replacing a VGG-16~\cite{simonyan2014} feature extractor with a modern one such as EfficientNet-B7~\cite{tan2019}. Instead, we seek to fairly compare the domain adaptation methods given the same exact networks and computational budget; only the loss varies.
\begin{table}
	\centering
	\caption{Hyper-parameter search space for domain adaptation methods.}
	\label{tab:office31-search-space}
	\begin{tabular}{lrrr}
		\toprule
        Hyper-Parameter 
		    & Lower
		    & Upper
		    & Prior
            \\
		\midrule
		Learning Rate
    		& $10^{-8}$
    		& $10^{-3}$	
    		& Log-Uniform
    		\\
	    Learning Rate Decay
    	    & $10^{-7}$
    		& $10^{-2}$
    		& Log-Uniform
    		\\
        Momentum 
            & $0.5$
    		& $0.99$
    		& Inv Log-Uniform
    		\\
    	Dropout
            & $0.1$
    		& $0.8$
    		& Uniform
    		\\
    	L2 Regularisation
            & $10^{-7}$
    		& $10^{-3}$
    		& Log-Uniform
    		\\
    	DA-CE Loss Ratio
            & $0.1$
    		& $0.99$	
    		& Uniform	
    		\\
    	$\Src$-$\Tgt$ CE Loss Ratio
            & $0$
    		& $1$
    		& Uniform
    		\\
    	No. Unfrozen Base-Layers
            & $0$
    		& $16$
    		& Uniform
    		\\
		\bottomrule
	\end{tabular}
\end{table}

\subsection{Office31}
The experimental setup for the Office31 datsets follows the procedure in
\cite{tzeng2015simultaneous, xu2019dsne, motiian2017ccsa}. 
First, a VGG16~\cite{simonyan2014} model with its convolutional layers pre-trained on ImageNet~\cite{Russakovsky2015} and randomly initialised dense layers of sizes 1024 and 128 is fine-tuned on all available source data. 
For this, we used a gradual-unfreeze procedure~\cite{howard2018universal}, unfreezing four pretrained layers each time the model converges.
The resulting model is denoted \emph{FT-Source}.
\begin{table*}
	\centering
	\caption{Macro average classification accuracy (\%) for Office-31 using a VGG16 network pretrained on ImageNet. The reported results are the mean and standard deviation across five runs. 
	}
	\label{tab:office31-results-vgg16}
	\resizebox{\textwidth}{!}{
	\begin{tabular}{lccccccc}
		\toprule
		    & $\concept{A} \rightarrow \concept{D}$
		    & $\concept{A} \rightarrow \concept{W}$
		    & $\concept{D} \rightarrow \concept{A}$
		    & $\concept{D} \rightarrow \concept{W}$
		    & $\concept{W} \rightarrow \concept{A}$
		    & $\concept{W} \rightarrow \concept{D}$
		    & Avg. \\
		\midrule
		FT-Source
		    & $66.6 \pm 3.0$
		    & $59.8 \pm 2.1$
		    & $42.8 \pm 5.2$
		    & $92.3 \pm 2.8$
		    & $44.0 \pm 0.7$
		    & $98.5 \pm 1.2$
		    & $67.4$ \\
		FT-Target 
		    & $71.4 \pm 2.0$
            & $74.0 \pm 4.9$
            & $56.2 \pm 3.6$
            & $95.9 \pm 1.2$
            & $50.2 \pm 2.6$
            & $99.1 \pm 0.8$
            & $74.5$ \\
		CCSA
		    & $84.8 \pm 2.1$	
		    & $87.5 \pm 1.5$	
		    & $\mathbf{66.5 \pm 1.9}$	
		    & $97.2 \pm 0.7$	
		    & $64.0 \pm 1.6$	
		    & $98.6 \pm 0.4$	
		    & $83.1$ \\
		\textit{d}-SNE
		    & $\mathbf{86.5	\pm 2.5}$ 
		    & $\mathbf{88.7 \pm	1.9}$ 
		    & $ 65.9 \pm 1.1 $ 
		    & $ 97.6 \pm 0.7 $ 
		    & $ 63.9 \pm 1.2 $ 
		    & $ 99.0 \pm 0.5 $
		    & $\mathbf{83.6}$ \\
		DAGE-LDA 
		    & $85.9 \pm 2.8$	
		    & $87.8 \pm 2.3$	
		    & $66.2 \pm 1.4$	
		    & $\mathbf{97.9 \pm 0.6}$	
		    & $\mathbf{64.2 \pm 1.2}$	
		    & $\mathbf{99.5 \pm 0.5}$	
		    & $\mathbf{83.6}$ \\

		\bottomrule
	\end{tabular}
	}
\end{table*}
Then, samples are selected randomly from source and target domains. 
Twenty samples per class are drawn from the source domain $\concept{A}$, whereas only eight samples per class are drawn from $\concept{D}$ and $\concept{W}$.
In all cases, three samples are chosen from the target domain and the remaining target samples are used for testing. 
The datasets are combined as the Cartesian product of the sampled source and target sets.
Each training sample is thus either a positive or negative pair of source and target samples depending on whether the corresponding labels are equal.
To limit the dataset size, negative pairs are randomly discarded to achieve a 3:1 ratio of negative to positive pairs. Using the resulting datasets and the FT-Source model as initial model weights, the domain adaptation is performed. 
As a simple baseline comparing the efficiency of domain adaptation to standard parameter-based transfer learning, we fine-tune a FT-Source model using all the target data. This is called \emph{FT-Target}. 
For each of the six transfers, the experiment is repeated five times.

To produce a fair comparison, we implemented DAGE-LDA as well as CCSA and $d$-SNE based on their publicly available code. 
An individual hyper-parameter search was conducted for each method and transfer prior to testing.
The results are shown in \cref{tab:office31-results-vgg16}. 
Following the practice in the source code of \cite{motiian2017ccsa} and \cite{xu2019dsne}, the test set was used to determine early stopping.
The reader should thus consider the published results of \cite{motiian2017ccsa}, \cite{xu2019dsne} as well as this paper as validation results.

It is clear that all the evaluated DA methods outperform the baseline (FT-Target) by a comfortable margin. In our experiments, $d$-SNE performed slightly better than CCSA, but less so than was concluded by  Xu et al.~\cite{xu2019dsne}. DAGE-LDA performs on par with $d$-SNE. 
For the $\concept{D} \leftrightarrow \concept{W}$ transfers, where the source and target data are very similar, our results generally beat the published results. We attribute this to our use of gradual unfreeze, which arguably succeeded in retaining useful knowledge from the source domain.
Despite our best efforts, we could not reproduce the results reported in \cite{motiian2017ccsa} and \cite{xu2019dsne} on the other transfers, neither with their source code or with our implementations. Moreover, we observed a significantly larger standard deviation of the results in our own experiments.

\subsection{MNIST $\rightarrow$ USPS}
Following the procedure in
\cite{fernando2014joint, motiian2017ccsa, xu2019dsne},
we sample 2,000 images from MNIST at random, as well as a small number of samples per class from USPS. Experiments using 1, 3, 5 and 7 target samples per class were conducted and each experiment was repeated 10 times.
The network architecture consists of two $5 \times 5$ convolutional layers with 6 and 16 filters, followed by two fully connected layers of sizes 120 and 84.
This is the same as reported by \cite{motiian2017ccsa} and \cite{xu2019dsne}, though their source code deviated from this.
Prior to testing their publicly available code, we made the minimal modifications necessary for the executed code to match the experimental description\footnote{\label{foot:ccsa-code-mod} \ifthenelse{\blind}{A link to the source code will be available upon acceptance}{Source code available at \url{https://github.com/lukashedegaard/CCSA}}}
\footnote{\label{foot:dsne-code-mod} \ifthenelse{\blind}{A link to the source code will be available upon acceptance}{Source code available at \url{https://github.com/sheikhomar/d-SNE}}}.
Moreover, an implementation bug in the CCSA codebase was fixed to ensure that the model parameters are reset between experiments.
In our own implementation, we used dropout, weight decay, and optionally batch norm with values found through an individual hyperparameter search for each method.

\begin{table}
	\centering
	\caption{Classification accuracy (\%) for MNIST $\rightarrow$ USPS with a varying number of target samples per class.
	}
	\label{tab:mnist-usps-results}
	\begin{tabular}{lcccc}
		\toprule
        Samples/class
		    & 1
		    & 3
		    & 5
		    & 7
		    \\
		\midrule
		CCSA
		    & $\mathbf{75.6 \pm 2.1}$
            & $\mathbf{85.0 \pm 1.4}$
            & $87.8 \pm 0.7	$
            & $89.1 \pm 0.7	$
            \\
		\textit{d}-SNE
		    & $69.0 \pm 1.7$
            & $80.4 \pm 1.7$
            & $86.1 \pm 0.9$
            & $87.7 \pm 0.9$
		    \\
		DAGE-LDA
		    & $67.0 \pm 1.9$
            & $82.7 \pm 1.7$
            & $\mathbf{89.0 \pm 0.8}$
            & $\mathbf{90.7 \pm 0.5}$
		    \\
		\bottomrule
	\end{tabular}
\end{table}

The results of our tests are shown in \cref{tab:mnist-usps-results}.
As in the Office 31 experiments, there is a gap between the reported results and what we reproduced.
The difference is especially large for the reproduction using the published codes for CCSA and $d$-SNE. For CCSA, the identified bug in the source code fully explains the discrepancy.
The $d$-SNE code did not contain an implementation of the network architecture described above. Instead LeNet$++$~\cite{wen2016conv} was implemented. As with CCSA, it was not possible to reproduce the reported results for $d$-SNE using their publicly available source code. We, among others, have raised issues on their GitHub testifying to this.
For a fair comparison with the other methods in our experiments, we modified the baseline architecture to match the others and performed hyperparameter optimisation with the same computational budget.

Interestingly, CCSA clearly outperformed the other methods for 1 and 3 target samples per class, whereas DAGE-LDA was best for 5 and 7. 
Contrary to our expectation, $d$-SNE performed worst for MNIST $\rightarrow$ USPS.

\subsection{Discussion of Experimental Setup}
In our effort to reproduce the results of CCSA and $d$-SNE, we saw a trend that was recently outlined in 
\cite{recht2018, recht2019, gencoglu2019hark},
namely that published results may not generalise as expected. 
Supporting this, is our observation that a validation set was not used in either of the published codes.
Furthermore, in the testing procedures, we observed that evaluation on the test set was performed after every epoch of training, with the end-result being the highest performing evaluation throughout training process. Hence, the test set was used as validation set in practice. 
We suspect that the results of many domain adaptation methods using these datasets is best viewed as the performance on a validation set. While this is not necessarily a problem if the result are used solely for comparison between methods using this same practice, it may be misleading readers who expect results to generalise.

%% file: content/05-conclusion.tex
\section{Conclusion} \label{sec:conclusion}
Domain adaptation methods help achieve better performance on tasks where data is scarce by leveraging larger related datasets to learn good feature representations. 
In this work, we treat domain adaptation as a dimensionality reduction problem and propose a novel use of Graph Embedding by integrating the trace-ratio objective as a loss in a deep neural network, that is trained end-to-end.
Using this Domain Adaptation Graph Embedding framework (DAGE), we test a simple LDA-inspired domain adaptation loss (DAGE-LDA) on standard benchmarking datasets and reevaluate CCSA and $d$-SNE, two the recent state-of-the-art methods, which can be seen as instantiations of DAGE.
Under identical experimental conditions, DAGE-LDA matches or beats the overall accuracy of both prior methods, highlighting the treatment of domain adaptation as a multi-view graph embedding problem.

%% file: content/06-acknowledgement.tex
\section{Acknowledgement}
Lukas Hedegaard and Alexandros Iosifidis acknowledge funding from the European Union’s Horizon 2020 research and innovation programme under grant agreement No 871449 (OpenDR). This publication reflects the authors’ views only. The European Commission is not responsible for any use that may be made of the information it contains.